\documentclass{article}
\usepackage{ijcai26}

\usepackage{times}
\usepackage{soul}
\usepackage{url}

\usepackage[hidelinks]{hyperref}

\usepackage[utf8]{inputenc}
\usepackage[small]{caption}
\usepackage{graphicx}
\usepackage{amsmath}
\usepackage{amsthm}
\usepackage{amssymb} 
\usepackage{booktabs}
\usepackage{algorithm}
\usepackage{algorithmic}
\usepackage[switch]{lineno}

\usepackage{subcaption} 
\usepackage{mathtools}

\title{How Well Do LLMs Perform on the Simplest Long-Chain Reasoning Tasks: \\An Empirical Study on the Equivalence Class Problem}

\author{
    Chun Zheng$^{1,*}$ \and Lianlong Wu$^{2,*}$ \and Bingqian Li$^1$ \and Lvting Liu$^1$ \and Yi Zhou$^1$
    \affiliations
    $^1$University of Science and Technology of China\\
    $^2$University of Oxford\\
    \emails
    zhengchun@mail.ustc.edu.cn, lianlong.wu@cs.ox.ac.uk, bqli315@gmail.com, yi\_zhou@ustc.edu.cn
}

\begin{document}

\maketitle
\begingroup
\renewcommand{\thefootnote}{\fnsymbol{footnote}}
\footnotetext[1]{These authors contributed equally.}
\endgroup

\begin{abstract}
Large Language Models (LLMs) have achieved great improvements in recent years. Nevertheless, it still remains unclear how good LLMs are for reasoning tasks, especially for long-chain ones. In this paper, we evaluate LLMs' performance on the simplest yet long-chain reasoning task, namely the Equivalence Class Problem (ECP), i.e., determining whether two variables are equal given a set of randomly generated equivalence relations. We consider both reasoning and non-reasoning representative LLMs over a large variety of problem instances, ranging over different numbers of variables, connectivity probabilities, prompts, and other factors. The experimental results show that non-reasoning LLMs fail ECP, while reasoning models are significantly better but still struggle to completely solve this problem. Interestingly, considering various connectivity probabilities with a fixed number of variables, we observe that, for non-reasoning models, the hardest problem instances coincide with the phase transition point of $\frac{\ln n}{n-1}$, suggesting the chaos of the problem; in contrast, for reasoning models, the hardest ones coincide with the biggest diameter, suggesting the reasoning difficulty of the problem.
\end{abstract}

\section{Introduction}

The evolution of Large Language Models (LLMs) has recently reached a watershed moment. The field currently exhibits a bifurcation into two coexisting paradigms: one consists of standard non-reasoning models, such as DeepSeek-V3~\cite{liu2024deepseek} and Qwen3-Max~\cite{qwen3max}, which rely on massive parameter counts and extensive training data for rapid pattern matching and text generation. The other comprises the recently emerging reasoning models, including Qwen3-MAX-thinking~\cite{qwen3max}, DeepSeek-V3.2-thinking~\cite{liu2025deepseek}, DeepSeek-R1~\cite{guo2025deepseek}, and Claude 4.5 Sonnet Thinking~\cite{claude_sonnet_4_5}. The distinguishing feature of the latter is the incorporation of Chain-of-Thought (CoT) mechanisms, enabling the generation of detailed reasoning traces before outputting a final answer. This ``System 2'' approach has demonstrated impressive performance gains on complex mathematical and programming benchmarks and is widely regarded as a crucial step toward more general artificial intelligence~\cite{wei2022chain,kojima2022large}.

However, despite the superior performance of both model paradigms on general leaderboards, the academic community lacks a deep understanding of their internal mechanisms and capability boundaries in long-chain reasoning scenarios. Current mainstream evaluation paradigms primarily rely on existing benchmarks such as AIME~\cite{maa_aime} or MATH~\cite{hendrycks2021measuring}. These paradigms suffer from two core limitations: First, data contamination and memorization effects, where models may simply memorize the mapping between questions and answers rather than truly mastering logical rules~\cite{mccoy2023embers}; Second, a lack of fine-grained control over complexity, as existing mathematical problems often entangle calculation, common sense, and logic, making it difficult to isolate a single variable and controllably test the impact of reasoning depth on performance~\cite{shojaee2025illusion,estermann2024puzzles}. A critical question remains unresolved: When semantic cues are stripped away, leaving only abstract logic, how do non-reasoning and reasoning models fundamentally differ? Are they performing genuine logical deduction, or merely engaging in complex probabilistic curve fitting?

To strictly evaluate logical capabilities, this study constructs a synthetic task environment based on the Equivalence Class Problem. This task is conceptually simple, requiring no external knowledge, yet it serves as a rigorous probe for long-chain reasoning by necessitating multi-step recursive deduction. Mathematically, the problem is equivalent to identifying connected components in a graph, where the model must infer global membership relationships solely through the transitivity of local equivalence links. This environment offers the unique advantage of isolating pure logical deduction from semantic interference, while being capable of generating problem instances that naturally span a diverse spectrum of inference depths.

Our empirical research is based on a comprehensive evaluation of representative models from both non-reasoning (e.g., DeepSeek-V3) and reasoning (e.g., DeepSeek-R1) paradigms. By systematically sweeping connectivity probabilities across varying problem scales (up to $n=144$) to cover the critical phase transition region, we reveal three key characteristics of current LLMs when processing abstract logic:

First, the error peaks diverge within the critical phase transition window (between $1/n$ and $\frac{\ln n}{n-1}$). DeepSeek-V3.2 (non-reasoning model) is overwhelmed by structural chaos, with its error rates peaking near the connectivity threshold $\frac{\ln n}{n-1}$. In contrast, DeepSeek-R1 (reasoning model) exhibits a depth-dependent failure: its error rates peak earlier in the transition, closer to $1/n$, precisely where the average inference depth reaches its maximum. This confirms that while standard models struggle with topological complexity, reasoning models are primarily constrained by the length of the deduction chain.

Second, non-reasoning models exhibit a severe limitation in handling multi-step dependencies. Our experiments with DeepSeek-V3.2 reveal that while the model performs robustly on single-hop deduction ($Depth=1$), it fails to generalize to longer chains. Specifically, the error rate increases sharply at $Depth=2$ and saturates to near-random levels by $Depth=3$. This indicates that standard Transformer architectures struggle to maintain logical coherence beyond immediate connections, regardless of model scale.

Third, reasoning models remain sensitive to the length of the reasoning chain. Although CoT-enhanced models like DeepSeek-R1 achieve significantly lower overall error rates compared to non-reasoning baselines, they have not achieved complete generalization. We observe that as the inference depth increases linearly, the error rate of these models grows exponentially. This indicates that while Chain-of-Thought mechanisms effectively extend the range of solvable steps, they do not fundamentally eliminate the accumulation of errors in recursive reasoning.

Furthermore, extensive ablations reveal that explicit rules and few-shot examples offer negligible gains, pointing to an execution bottleneck. While graph-theoretic framing significantly outperforms abstract logic, neither context isolation nor multi-path sampling (Pass@5) can prevent the fundamental reasoning collapse at the critical phase transition.

The main contributions of this paper are as follows:
\begin{itemize}
    \item We construct a controlled experimental platform based on equivalence class problem (ECP), providing a benchmark for assessing the pure logical reasoning boundaries of LLMs.
    \item We quantitatively characterize the failure of non-reasoning models in multi-step deduction, revealing their specific inability to perform reliable reasoning beyond single-hop tasks despite high accuracy on direct queries.
    \item We demonstrate that even in state-of-the-art reasoning models, long-chain capabilities remain subject to exponential error growth, refuting the optimistic assumption that ``reasoning models have solved logical generalization''.
    \item We provide empirical evidence that static prompt engineering is insufficient to overcome structural reasoning barriers, highlighting the necessity of System 2-like active computation.
\end{itemize}

\section{Related Work}

\paragraph{Evolution of LLMs and Reasoning Paradigms}
The rapid development of Large Language Models (LLMs) is widely regarded as a critical path toward Artificial General Intelligence (AGI), where the emergence of intelligence and reasoning capabilities has remained a core focus of research~\cite{mccoy2023embers,nezhurina2024alice,hu2024minictx}. Early research on Chain-of-Thought (CoT) found that guiding models to output intermediate reasoning steps before generating the final answer~\cite{wei2022chain,zhou2022teaching,kojima2022large} significantly improves accuracy on complex tasks. This discovery influenced the inception of ``reasoning models'': through reinforcement learning (RL) fine-tuning, models learn to explicitly generate implicit thinking processes within special \texttt{<think>} tokens~\cite{guo2025deepseek}. This paradigm shift has not only significantly boosted performance across various benchmarks but also endowed models with certain generalization capabilities~\cite{ma2025general,xu2025towards}. In the latest frontier advancements (e.g., DeepSeek-V3.2~\cite{liu2025deepseek}, Qwen3-Max~\cite{qwen3max}), this ``thinking'' capability is becoming increasingly modularized, allowing users to dynamically toggle it during the inference phase based on task requirements, marking a new trend toward the flexible allocation of reasoning computational resources.

\paragraph{Nature and Limitations of Reasoning Capabilities}
As reasoning models become ubiquitous, the academic community has begun to deeply dissect their internal mechanisms and capability boundaries~\cite{chen2025reasoning,li2025llms}. On one hand, microscopic analysis of reasoning traces has revealed complex behavioral patterns, ranging from self-reflection to ``overthinking''~\cite{chen2024not,marjanovic2025deepseek,sui2025stop}. On the other hand, controversy remains regarding whether RL truly endows models with ``novel'' reasoning capabilities. Some studies point out that when controlling for computational cost (e.g., pass@k testing), the performance difference between RL-trained reasoning models and base models is often negligible in many tasks, suggesting that RL may be eliciting existing capabilities rather than creating new ones~\cite{yue2025does}. Furthermore, research focusing on classic algorithmic puzzles like the Tower of Hanoi and Checkers finds that as problem complexity increases, both reasoning and non-reasoning models suffer from performance collapse~\cite{shojaee2025illusion}. Unlike these puzzles, which focus on testing ``algorithmic execution'' skills, this study constructs abstract Equivalence Class Problems aimed at stripping away semantic interference to purely evaluate the reasoning robustness of models in long-chain logical contexts.

\begin{figure*}[t!]
    \centering
    \includegraphics[width=1.0\textwidth]{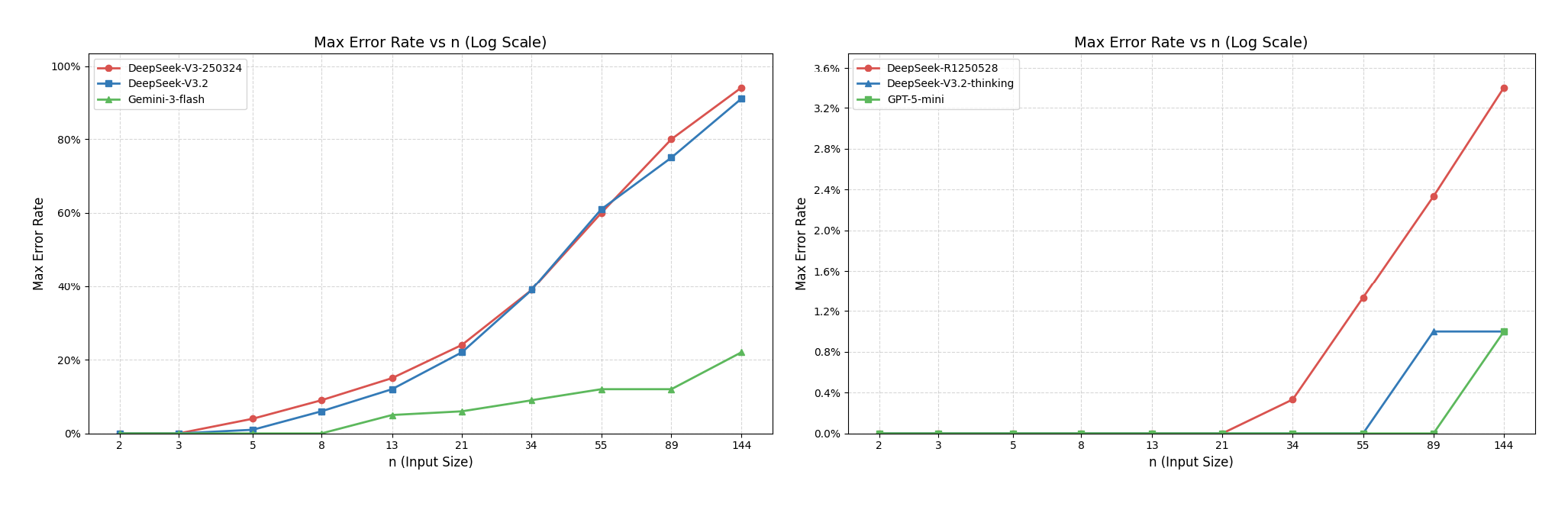}
    \caption{\textbf{Scaling Limit.} Maximum error rate vs. problem size ($n$). The left panel reports non-reasoning models, whose peak error rises sharply as the variable set grows. The right panel reports reasoning models, which substantially reduce the error scale but still expose nonzero failures under larger instances. Each point is the maximum error observed over the swept probability range for that problem size.}
    \label{fig:scaling_max}
\end{figure*}

\paragraph{Controllable Evaluation Environments}
To overcome the limitations of traditional static benchmarks (such as data contamination and unquantifiable difficulty), researchers are increasingly turning to ``Controllable Evaluation Environments''~\cite{estermann2024puzzles,valmeekam2022large,gui2025logicgame,shojaee2025illusion}. The core advantage of such environments lies in their ability to quantitatively map model performance boundaries by finely tuning problem complexity via parameterization, while maintaining consistent logical structures (such as rule definitions). Following this methodology, this study designs an experimental framework based on equivalence classes. This framework allows for the systematic control of inference depth and probability density, providing a controlled testbed for comparative analysis of behavioral differences between reasoning and non-reasoning models in long-chain reasoning tasks.

\section{Methodology}
\label{sec:methodology}

In this section, we formalize the evaluation of long-chain reasoning capabilities as an Equivalence Class Partitioning problem rooted in graph theory. This controlled environment allows us to precisely modulate the reasoning width and depth by adjusting the problem size and connectivity probability.

\subsection{Task Formalization}

We model the equivalence class problem using a random graph framework. Given the number of variables $n$ and a connectivity probability $p$, the task is defined over a graph $G = (V, E)$:

\paragraph{Variables.} Let $V = \{v_1, v_2, \dots, v_n\}$ be a set of $n$ distinct abstract variables (e.g., $\{a_1, a_2, \dots, a_n\}$).

\paragraph{Equivalence Relations.} We employ the Erd\H{o}s-R\'enyi model $G(n, p)$ to generate the underlying logical structure~\cite{gilbert1959random,erdds1959random}. For every pair of distinct variables $(v_i, v_j)$, a direct equivalence relation (represented as an undirected edge) is established independently with probability $p$. Let $E$ denote the set of all generated direct relations. In this setting, $p$ also corresponds to the expected graph density, since the expected fraction of realized edges among all $\binom{n}{2}$ possible undirected edges is $p$.

\subsection{Probing Mechanism: Pairwise Queries}
\label{sec:probing}

While the theoretical objective is to compute the full partition $\mathcal{P}$, evaluating the generative output of full sets can be prone to parsing errors and format hallucinations. To rigorously quantify the model's reasoning accuracy, we simplify the output format into a binary classification task via \textbf{Pairwise Queries}.

For a generated graph $G=(V, E)$, we construct a query set $Q = \{(u_i, v_i)\}_{i=1}^k$. For each pair $(u, v)$, the model is presented with the edge list $E$ and asked:
\begin{center}
    \textit{``Based on the relations above, is $u$ equivalent to $v$?"}
\end{center}
The ground truth $y(u, v)$ is determined by the graph connectivity:
\begin{equation}
    y(u, v) = \mathbb{I}[u, v \in \text{SameComponent}(G)]
\end{equation}
where $\mathbb{I}[\cdot]$ is the indicator function. The model is required to output a binary response (Yes/No). This setup allows us to explicitly test whether the model has successfully captured the transitive closure of the equivalence relations.

\begin{figure*}[t!]
    \centering
    \begin{subfigure}{0.48\textwidth}
        \centering
        \includegraphics[width=\linewidth, height=5cm, keepaspectratio]{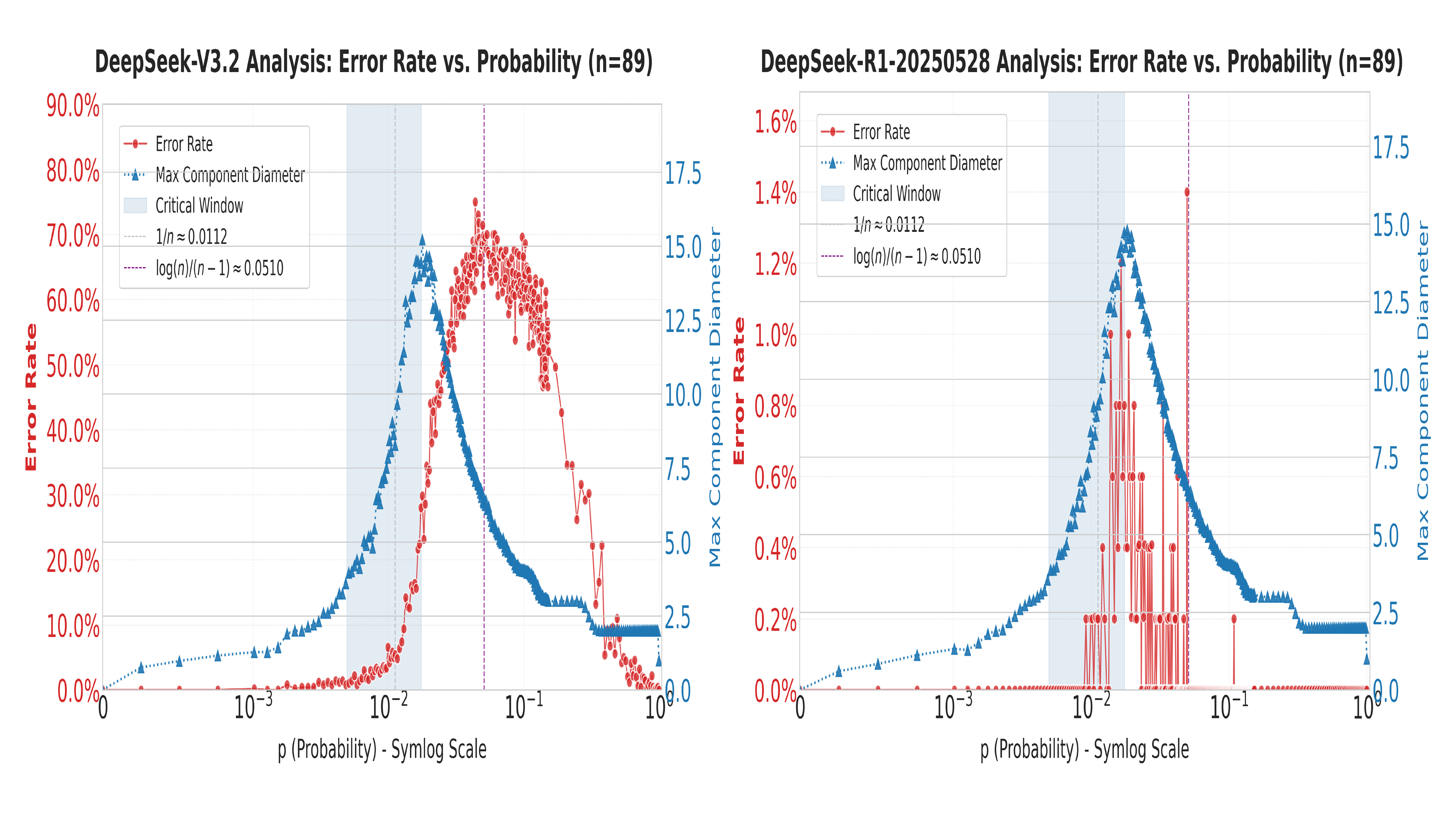}
        \caption{Error Rate vs. Probability ($n=89$)}
        \label{fig:89_p_err}
    \end{subfigure}
    \hfill
    \begin{subfigure}{0.48\textwidth}
        \centering
        \includegraphics[width=\linewidth, height=5cm, keepaspectratio]{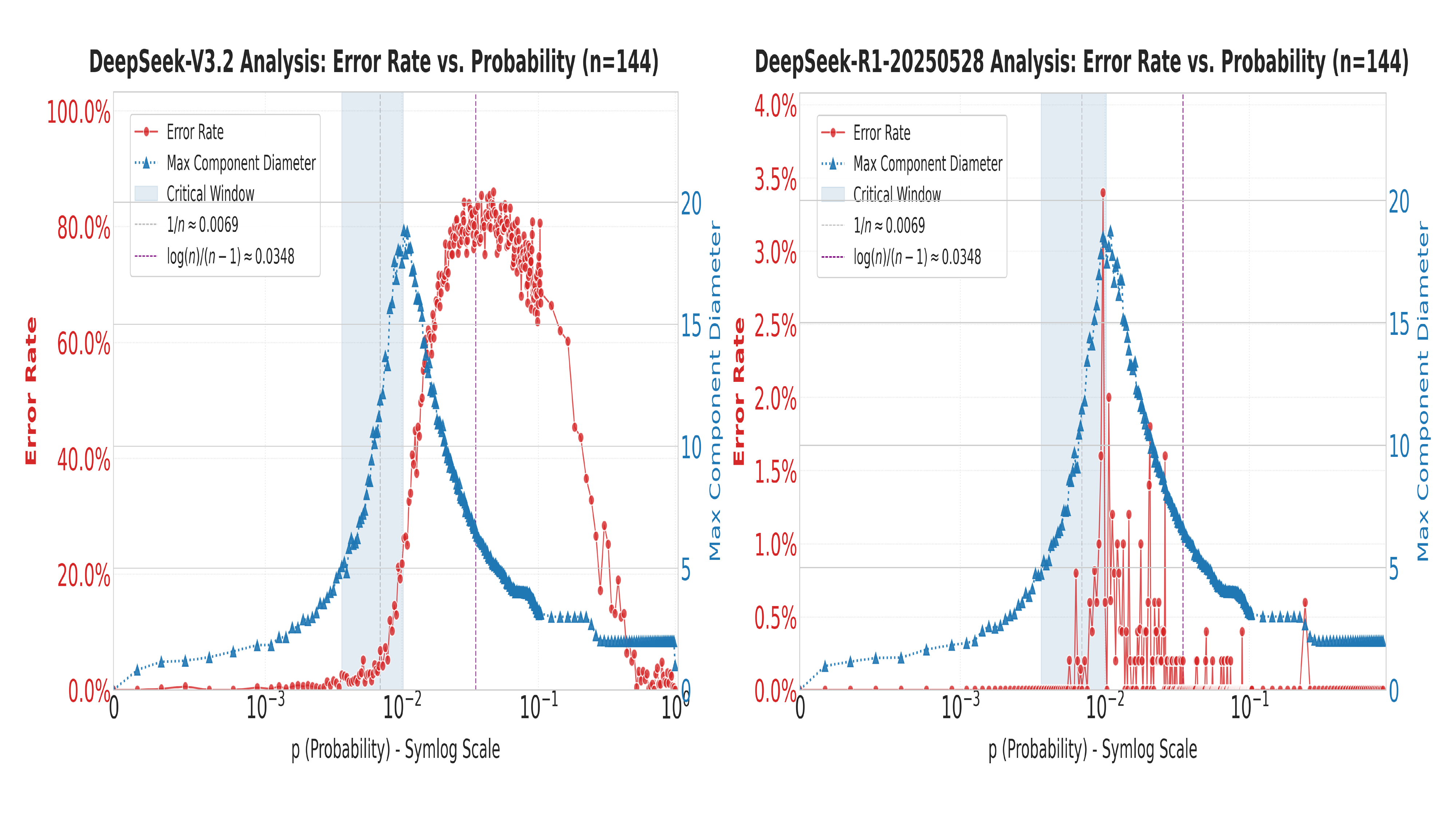}
        \caption{Error Rate vs. Probability ($n=144$)}
        \label{fig:144_p_err}
    \end{subfigure}
    
    \caption{\textbf{Error Rate vs. Probability Analysis ($n=89$ and $n=144$).} The x-axis is the edge probability $p$ used to sample equivalence relations; the y-axis is query error rate. The two subfigures compare DeepSeek-V3.2 and DeepSeek-R1-250528 at $n=89$ and $n=144$. Vertical reference markers indicate the sparse-chain regime around $1/n$ and the connectivity threshold around $\ln n/(n-1)$, where the model failure modes concentrate.}
    \label{fig:prob_analysis}
\end{figure*}

\subsection{Evaluation Protocol}
\label{sec:protocol}

To systematically map the capability boundaries of LLMs, we implement a dense sweeping strategy across the probability space:

\begin{enumerate}
    \item \textbf{Probability Sweep:} We traverse the connectivity probability $p$ from $0$ to $1$. The sampling density is increased near the critical phase transition points ($1/n$ and $\ln n/(n-1)$) to capture fine-grained behavioral changes~\cite{erd6s1960evolution}.
    \item \textbf{Instance Sampling:} For each probability point $p$, we randomly generate $N_{graphs}$ distinct graph instances to ensure statistical significance.
    \item \textbf{Query Sampling:} For each graph instance, we sample $k=10$ random pairs of variables $(u, v)$ as test queries. 
    \item \textbf{Metric:} The performance is evaluated using the \textbf{Error Rate}, defined as the fraction of queries where the model's prediction differs from the ground truth $y(u, v)$.
\end{enumerate}

This protocol ensures that our evaluation covers the full spectrum of complexity, from sparse, disconnected graphs to dense, fully connected networks.

\section{Experiments \& Results}

\subsection{Experimental Setup}
\label{sec:setup}

We conduct experiments using a diverse set of Large Language Models to represent distinct paradigms. For the \textbf{non-reasoning paradigm}, we evaluate \textbf{DeepSeek-V3} (including snapshots V3-250324 and V3.2)~\cite{liu2025deepseek,liu2024deepseek} alongside \textbf{Gemini-3-Flash}~\cite{Gemini3flash} as a strong baseline. For the \textbf{reasoning paradigm}, we employ \textbf{DeepSeek-R1}, \textbf{DeepSeek-R1-250528}~\cite{guo2025deepseek},  \textbf{DeepSeek-V3.2-thinking}~\cite{liu2025deepseek} and \textbf{GPT-5-mini}~\cite{GPT5mini}. To accommodate the extensive edge list descriptions in dense graphs, the context window is set to \textbf{60,000 tokens} for all tests. Following the protocol in Section~\ref{sec:protocol}, we primarily report the \textbf{Average Error Rate}, calculated by aggregating results from 10 randomly sampled pairwise queries across multiple independent graph instances for each configuration. Unless otherwise stated, we employ \textbf{Prompt 1} with temperature $T=0.01$ (pass@1). The detailed specifications for Prompt 1, 2, and 3 are provided in the ablation analysis (see \hyperref[par:rule_impact]{\textit{4.4.1}}).

\subsection{Main Results: Scaling Limits and Reasoning Collapse}
\label{sec:main_results}

We analyze the fundamental reasoning capabilities of the models across varying problem scales ($n$) and inference depths. The experimental results, encompassing consecutive Fibonacci scales ($n=89, 144$), reveal consistent failure patterns that persist regardless of problem size.

\begin{figure*}[t!]
    \centering
    \begin{subfigure}{0.48\textwidth}
        \centering
        \includegraphics[width=\linewidth, height=5cm, keepaspectratio]{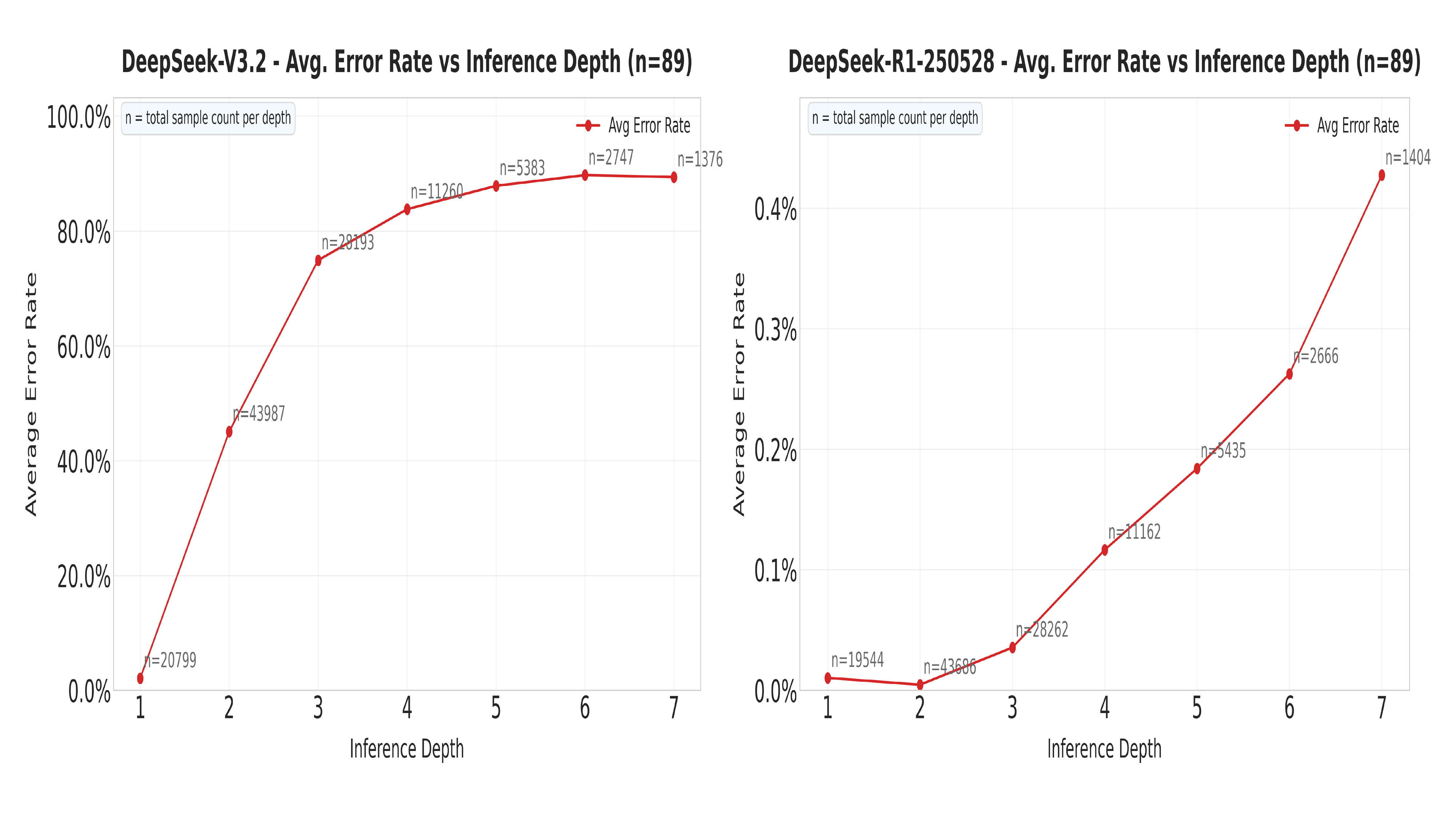}
        \caption{Error Rate vs. Inference Depth ($n=89$)}
        \label{fig:89_depth_err}
    \end{subfigure}
    \hfill
    \begin{subfigure}{0.48\textwidth}
        \centering
        \includegraphics[width=\linewidth, height=5cm, keepaspectratio]{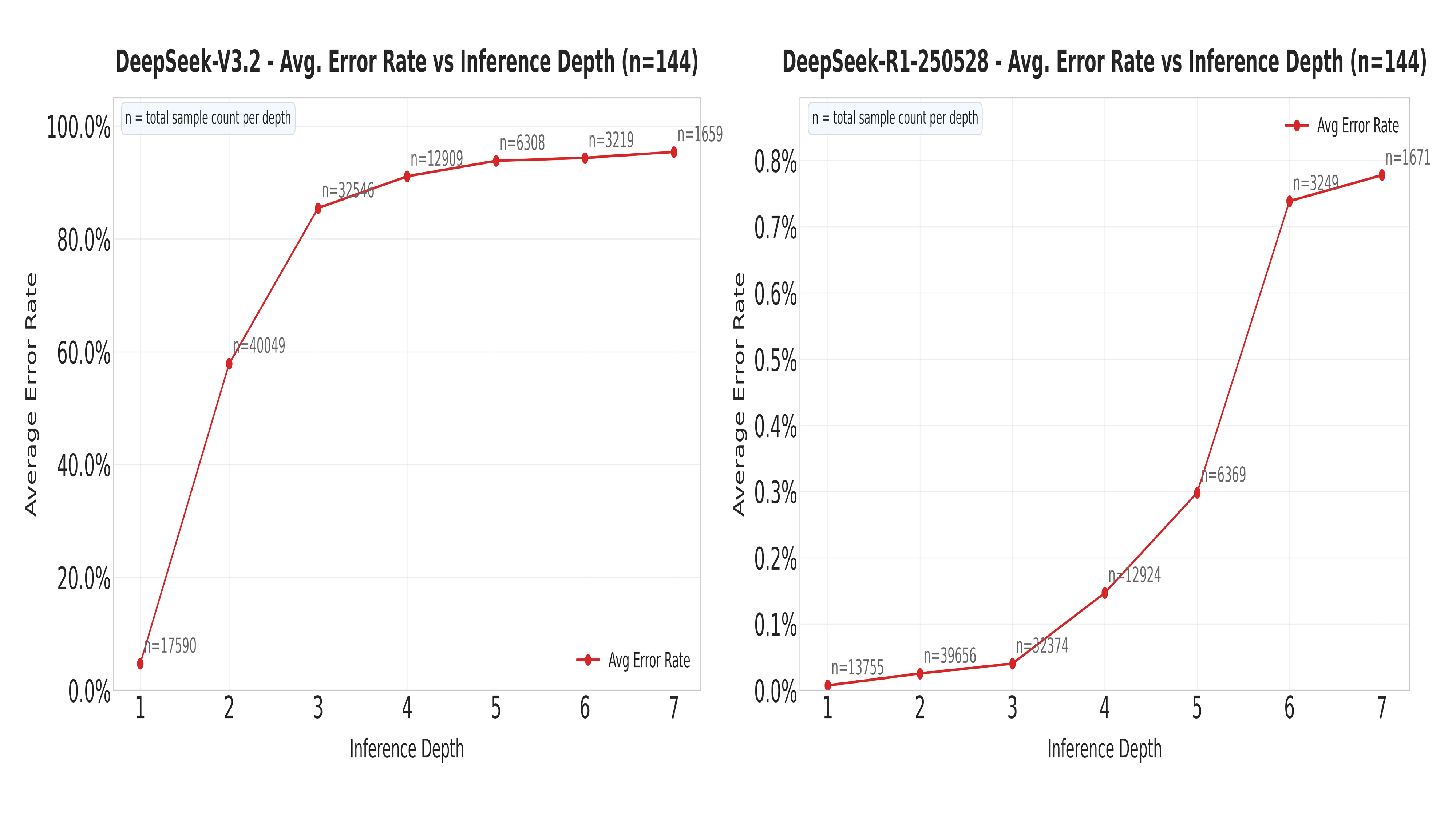} 
        \caption{Error Rate vs. Inference Depth ($n=144$)}
        \label{fig:144_depth_err}
    \end{subfigure}
    
    \caption{\textbf{Inference Depth Analysis ($n=89$ and $n=144$).} The x-axis is the shortest-path inference depth between the queried variables, and the y-axis is the corresponding average error rate. DeepSeek-V3.2 collapses immediately beyond single-hop inference (Depth $\ge$ 2), whereas DeepSeek-R1-250528 keeps errors low but still shows depth-dependent accumulation.}
    \label{fig:depth_analysis}
\end{figure*}

\paragraph{Scaling with Problem Size.}
As shown in Figure~\ref{fig:scaling_max}, the \textbf{Maximum Error Rate} (i.e., the peak error observed across the full probability spectrum $p$) exhibits a clear upward trend as the problem size increases from $n=2$ to $n=144$, but with vastly different magnitudes across paradigms.

\begin{itemize}
    \item \textbf{Non-Reasoning Collapse:} For \textbf{DeepSeek-V3} (Figure~\ref{fig:scaling_max} Left), the performance degradation is catastrophic. The error rate rises sharply, exceeding \textbf{90\%} at $n=144$ for both V3 and V3.2 variants. This indicates a near-total breakdown in logical consistency as the search space expands. Interestingly, \textbf{Gemini-3-Flash} exhibits significantly better robustness, capping at approximately \textbf{22\%} error at $n=144$, suggesting that while architecture matters, specific training recipes can mitigate some (but not all) logical deficits in non-reasoning models.
    
    \item \textbf{Reasoning Robustness:} Reasoning models (Figure~\ref{fig:scaling_max} Right) universally suppress error scales, though with evolving stability profiles. The latest architectures, \textbf{DeepSeek-V3.2-thinking} and \textbf{GPT-5-mini}, demonstrate near-perfect scalability, both maintaining an exceptional error rate of $\sim$\textbf{1.0\%} even at $n=144$. In comparison, the pioneer \textbf{DeepSeek-R1-250528}, while still highly robust, exhibits a slight linear rise in errors after $n=34$ to reach $\sim$\textbf{3.4\%}. This progression indicates that recent optimizations in ``thinking'' models have effectively mitigated complexity leakage, further compressing the error margin beyond the initial capabilities of R1.
\end{itemize}

\paragraph{Phase Transitions and Probability ($n=89$ vs. $n=144$).}
Figure~\ref{fig:prob_analysis} compares the error distributions across the probability spectrum $p$. A universal phenomenon is observed across both problem sizes:
\begin{itemize}
    \item Error rates are not uniformly distributed but peak precisely near the theoretical phase transition region, specifically aligning with the emergence of the giant component (between $1/n$ and $\ln n/(n-1)$).
    \item For \textbf{DeepSeek-V3.2}, the error distribution is broad and catastrophic, covering the entire ``super-critical" phase. As shown in Figure~\ref{fig:prob_analysis} (Left), the error rate rapidly saturates to $>80\%$ immediately after the critical threshold ($\log(n)/(n-1)$), indicating a total failure to navigate complex logical topologies.
    \item For \textbf{DeepSeek-R1-250528}, the error landscape is fundamentally different. While distinct error spikes still appear near the critical transition point (where logical complexity is maximized), the magnitude is suppressed by orders of magnitude (peaking at $<3.5\%$ compared to V3.2's $>85\%$). This confirms that R1 effectively handles the vast majority of transitive deductions, faltering only at the chaotic edge of connectivity.
\end{itemize}

\begin{figure*}[t!]
    \centering
    \includegraphics[width=1.0\textwidth]{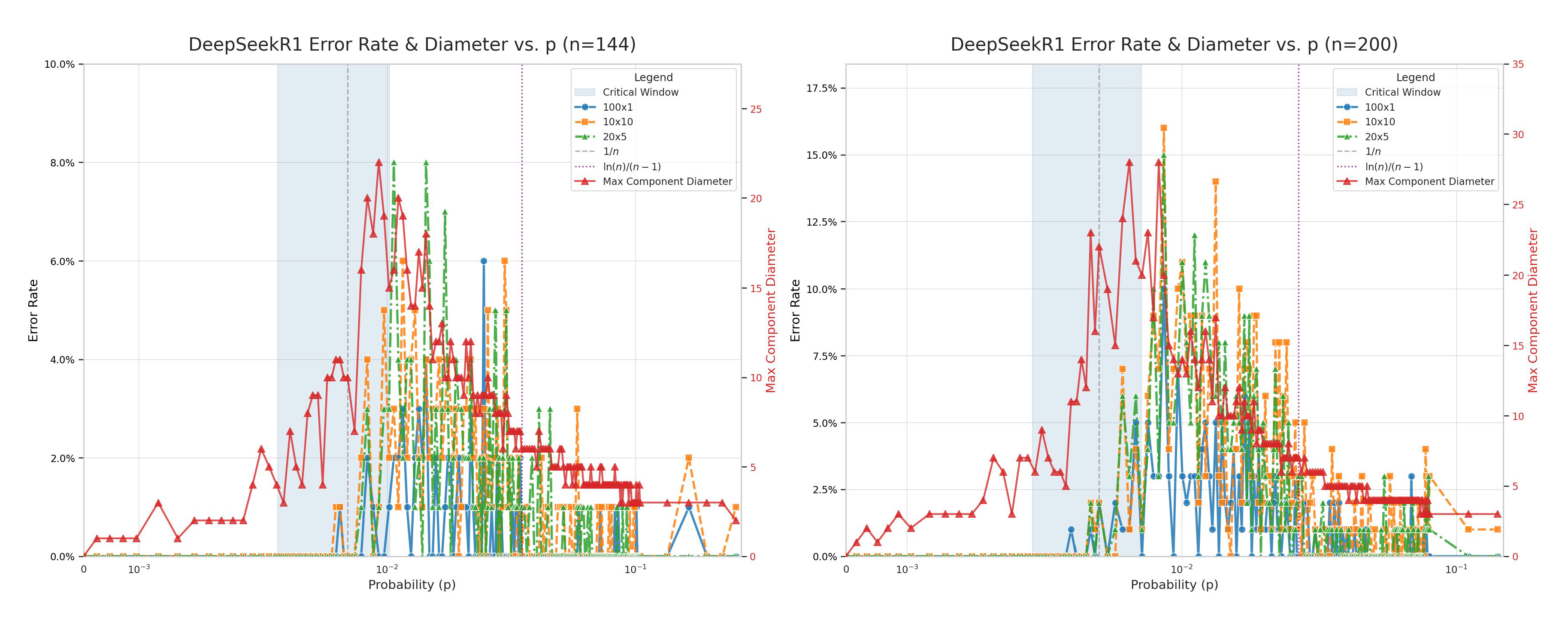}
    \caption{\textbf{Contextual Robustness of DeepSeek-R1 ($n=144$ and $n=200$).} 
    Comparison of error rates under different batching strategies: $100 \times 1$ (1 query per graph), $20 \times 5$, and $10 \times 10$. 
    \textbf{Left ($n=144$):} The $10 \times 10$ setting (Orange) shows higher error variance compared to the stable $100 \times 1$ setting (Blue).
    \textbf{Right ($n=200$):} As complexity increases, the gap widens. Notably, even the cleanest $100 \times 1$ setting fails to solve the problem near the phase transition, indicating fundamental reasoning limitations beyond mere context window constraints.}
    \label{fig:longcontext}
\end{figure*}

\paragraph{The Inference Horizon Barrier.}
The relationship between average error rate and inference depth (Figure~\ref{fig:depth_analysis}) delineates a sharp boundary between the two model paradigms, particularly at the scale of $n=144$.

\begin{itemize}
    \item \textbf{DeepSeek-V3.2 (Non-Reasoning):} The updated results reveal a striking \textbf{``One-Hop Limit''}. As illustrated in Figure~\ref{fig:depth_analysis} (Middle Left), V3.2 achieves a surprisingly low error rate of $\sim$4.4\% at Depth=1, suggesting robust performance on direct premise retrieval. However, logical coherence collapses immediately upon extending the chain: the error rate skyrockets to $\sim$58\% at Depth=2 and hits a saturation plateau of $\sim$85\% by Depth=3. This steep ``Reasoning Wall'' confirms that despite model updates, the non-reasoning paradigm remains fundamentally incapable of reliable multi-step deduction, treating $A \to B \to C$ as disjointed facts rather than a connected chain.

    \item \textbf{DeepSeek-R1-250528 (Reasoning):} In contrast, R1 demonstrates remarkable robustness. The error rate remains negligible ($\sim 0.01\%$) for shallow depths and maintains a strictly controlled profile even at deeper levels. However, the \textbf{exponential trend} persists: errors grow geometrically from $\sim 0.04\%$ at Depth 3 to $\sim 0.78\%$ at Depth 7. While the absolute error remains low ($<1\%$), the trajectory indicates that even sophisticated reasoning models are subject to multiplicative error accumulation as the reasoning chain lengthens.
\end{itemize}

\subsection{Contextual Robustness in Reasoning Models}
\label{sec:robustness}

While the reasoning model DeepSeek-R1 demonstrates superior performance compared to non-reasoning baselines, a critical question remains: is its reasoning capability robust to context saturation? To investigate this, we designed a ``Contextual Interference" experiment. We kept the total number of evaluation queries constant (100 queries per probability point) but varied the distribution of queries per graph instance. Specifically, we compared three batching strategies:
\begin{itemize}
    \item \textbf{Low-Interference ($100 \times 1$):} 100 graphs, each followed by a single query. This represents the purest reasoning environment with minimal context history.
    \item \textbf{Medium-Interference ($20 \times 5$):} 20 graphs, each followed by 5 consecutive queries.
    \item \textbf{High-Interference ($10 \times 10$):} 10 graphs, each followed by 10 consecutive queries.
\end{itemize}

As illustrated in Figure~\ref{fig:longcontext}, we observe two distinct phenomena across both $n=144$ and $n=200$:

\paragraph{1. The Attention Dilution Effect.}
There is a clear performance hierarchy: $100 \times 1$ (Blue) outperforms $20 \times 5$ (Green), which in turn outperforms $10 \times 10$ (Orange). The error rate increases significantly when the model is required to resolve multiple queries within a single context window. This suggests that despite the Chain-of-Thought mechanism, reasoning models suffer from \textit{attention dilution} or state-tracking degradation when maintaining the logical state over an extended interaction turn.

\paragraph{2. Intrinsic Logical Limits.}
Crucially, even in the most favorable setting ($100 \times 1$), the error rate \textbf{does not converge to zero}. As shown in the $n=200$ subplot (Right), the $100 \times 1$ configuration still exhibits substantial error spikes (reaching $>$10\%) near the phase transition region. This confirms that the observed failures are not solely artifacts of context length limitations or distraction, but stem from intrinsic deficits in the model's ability to generalize equivalence class rules at scale.

\subsection{Ablation Studies on Non-Reasoning Models}
\label{sec:ablation}

Given the structural limitations observed in DeepSeek-V3.2 (Section~\ref{sec:main_results}), we performed extensive ablation studies to determine whether external prompting strategies or decoding adjustments could mitigate these deficits. All experiments were conducted at the critical scale of $n=144$.

\paragraph{1. Impact of Rule Specification in Prompts.}\label{par:rule_impact}
We evaluated whether the ``Reasoning Wall" could be breached by providing explicit logical scaffolding.
\begin{itemize}
    \item \textbf{Prompt 1 (Baseline):} Full instructions including explicit transitivity/reflexivity definitions and role-playing (Mathematician).
    \item \textbf{Prompt 2:} Removed explicit rule definitions (relying on the model's internal concept of ``equivalence").
    \item \textbf{Prompt 3:} Removed both rule definitions and role-playing.
\end{itemize}

\begin{figure*}[t!]
    \centering
    \begin{subfigure}{0.48\textwidth}
        \centering
        \includegraphics[width=\linewidth, height=4.0cm, keepaspectratio]{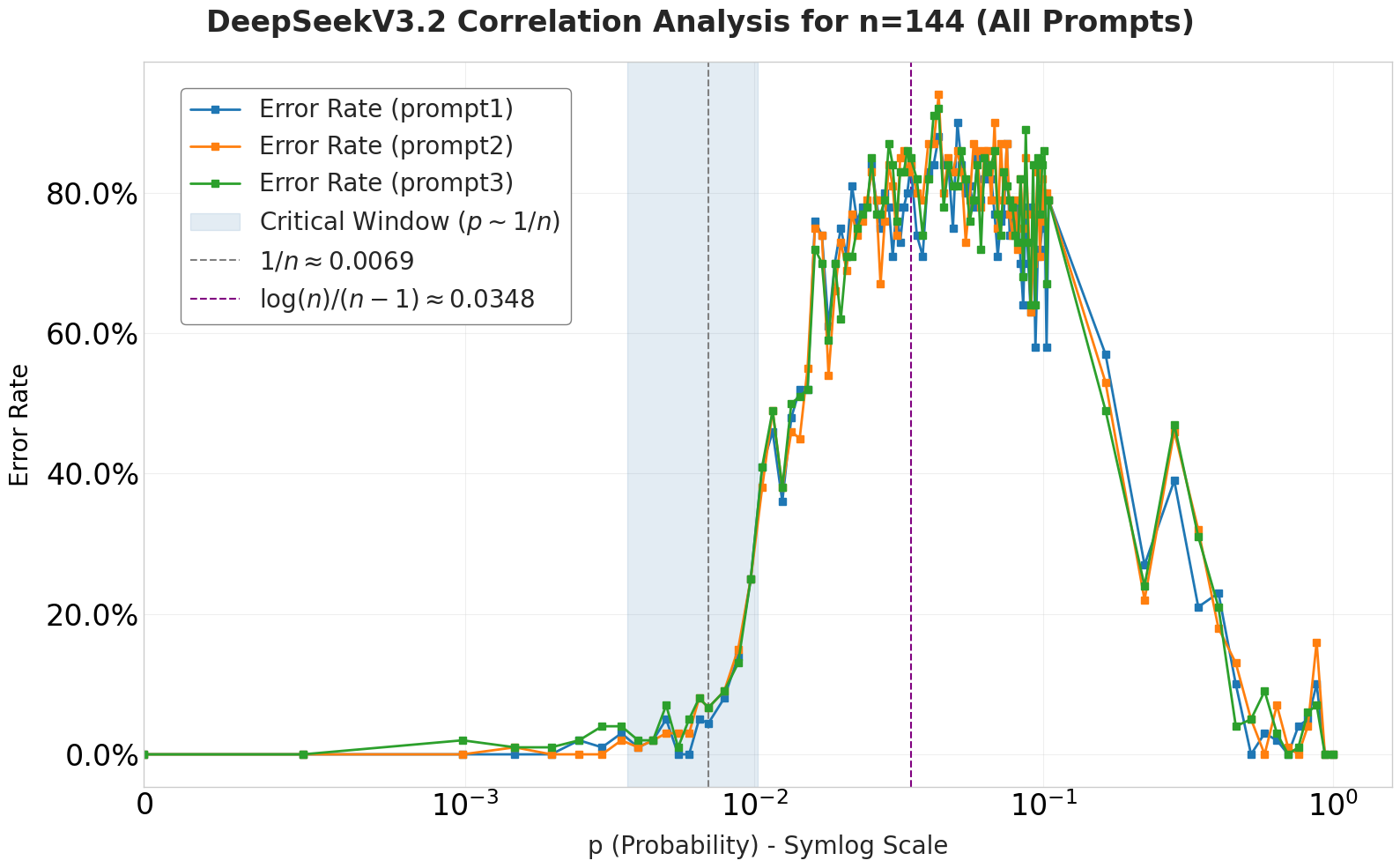}
        \caption{Rule Specification (P1 vs P2 vs P3)}
        \label{fig:ablation_rules}
    \end{subfigure}
    \hfill
    \begin{subfigure}{0.48\textwidth}
        \centering
        \includegraphics[width=\linewidth, height=4.0cm, keepaspectratio]{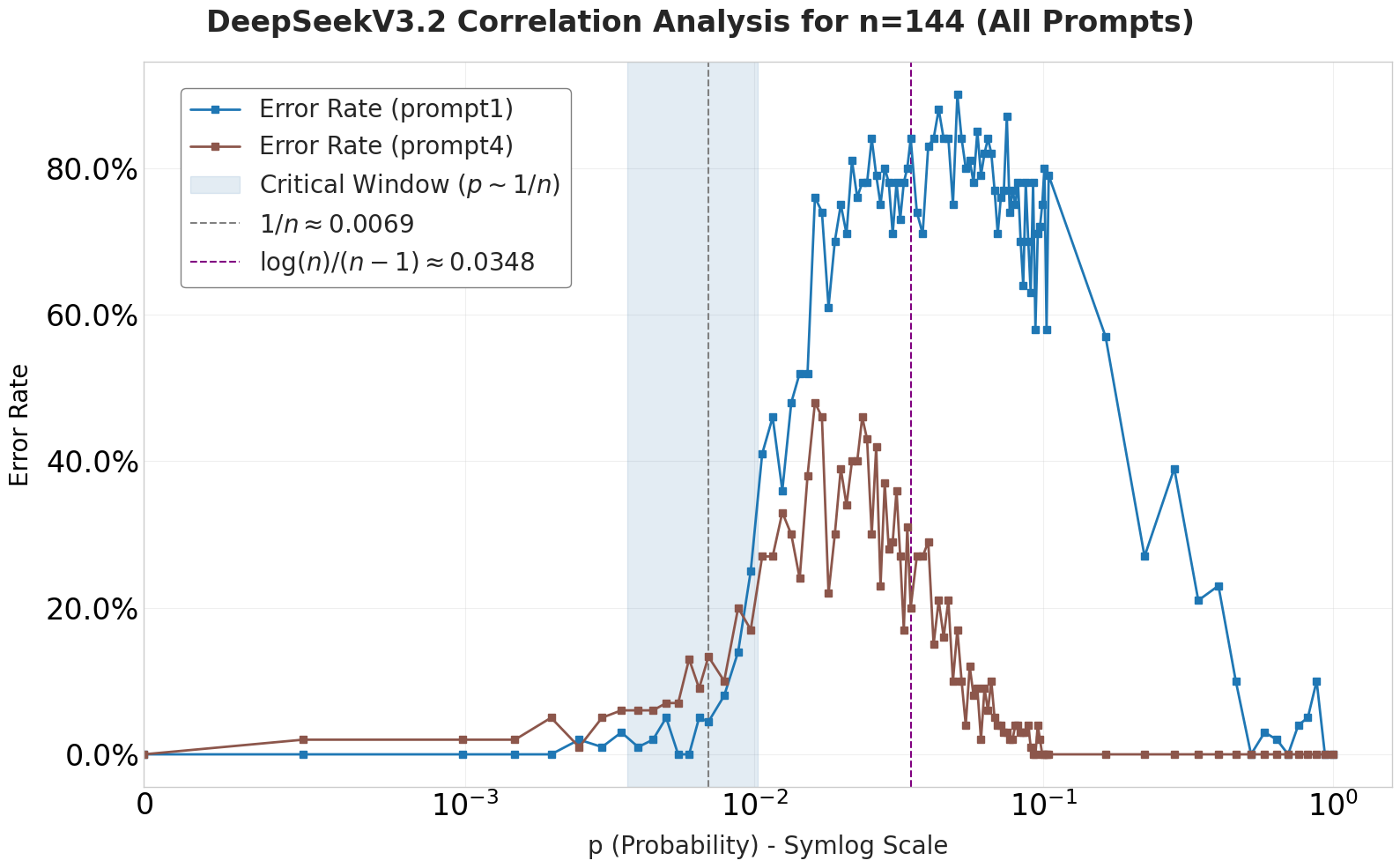}
        \caption{Domain Framing (Logic P1 vs Graph P4)}
        \label{fig:ablation_domain}
    \end{subfigure}
    
    \vspace{0.3cm}
    
    \begin{subfigure}{0.48\textwidth}
        \centering
        \includegraphics[width=\linewidth, height=4.0cm, keepaspectratio]{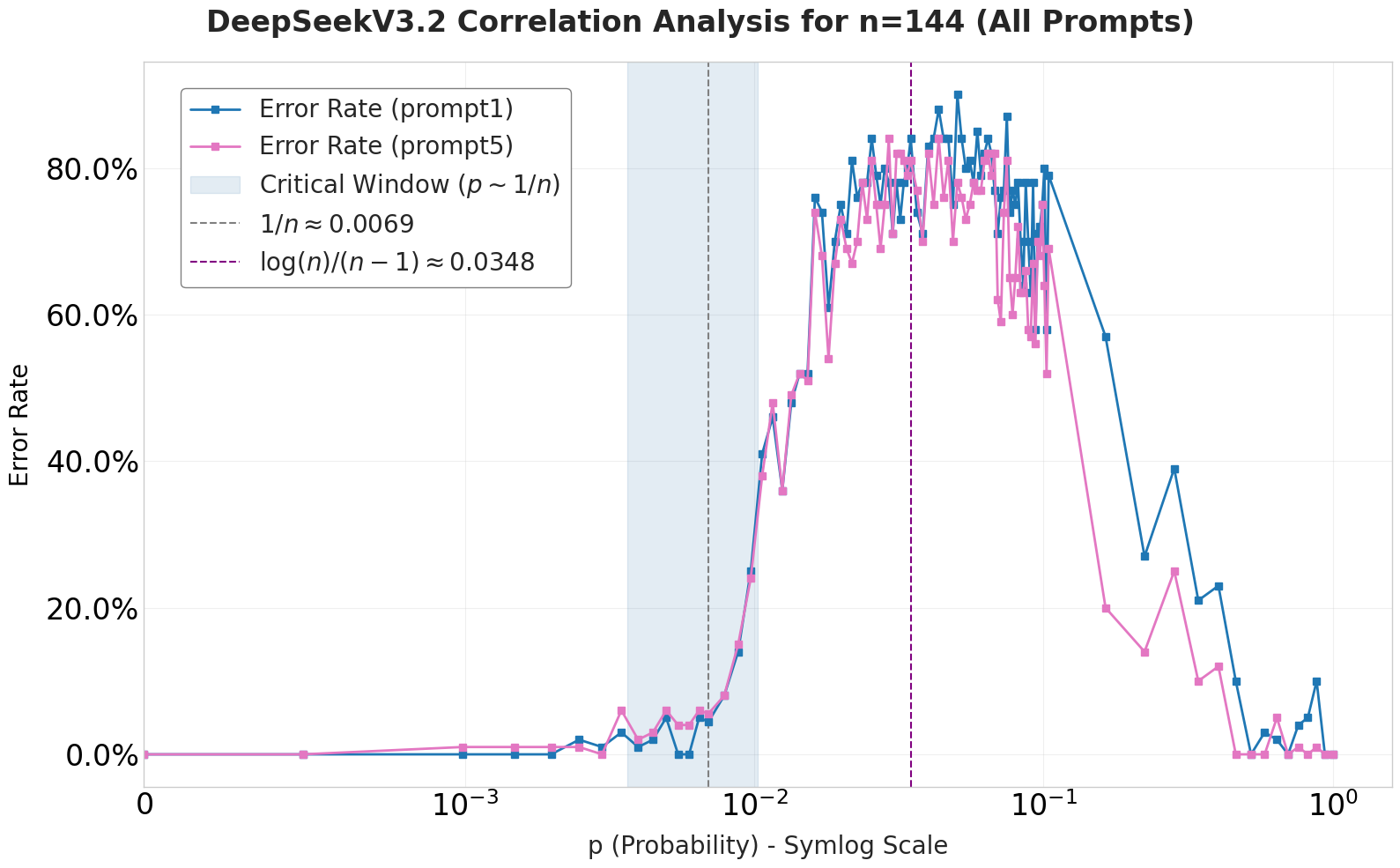}
        \caption{In-Context Learning (Zero-shot P1 vs Few-shot P5)}
        \label{fig:ablation_icl}
    \end{subfigure}
    \hfill
    \begin{subfigure}{0.48\textwidth}
        \centering
        \includegraphics[width=\linewidth, height=4.0cm, keepaspectratio]{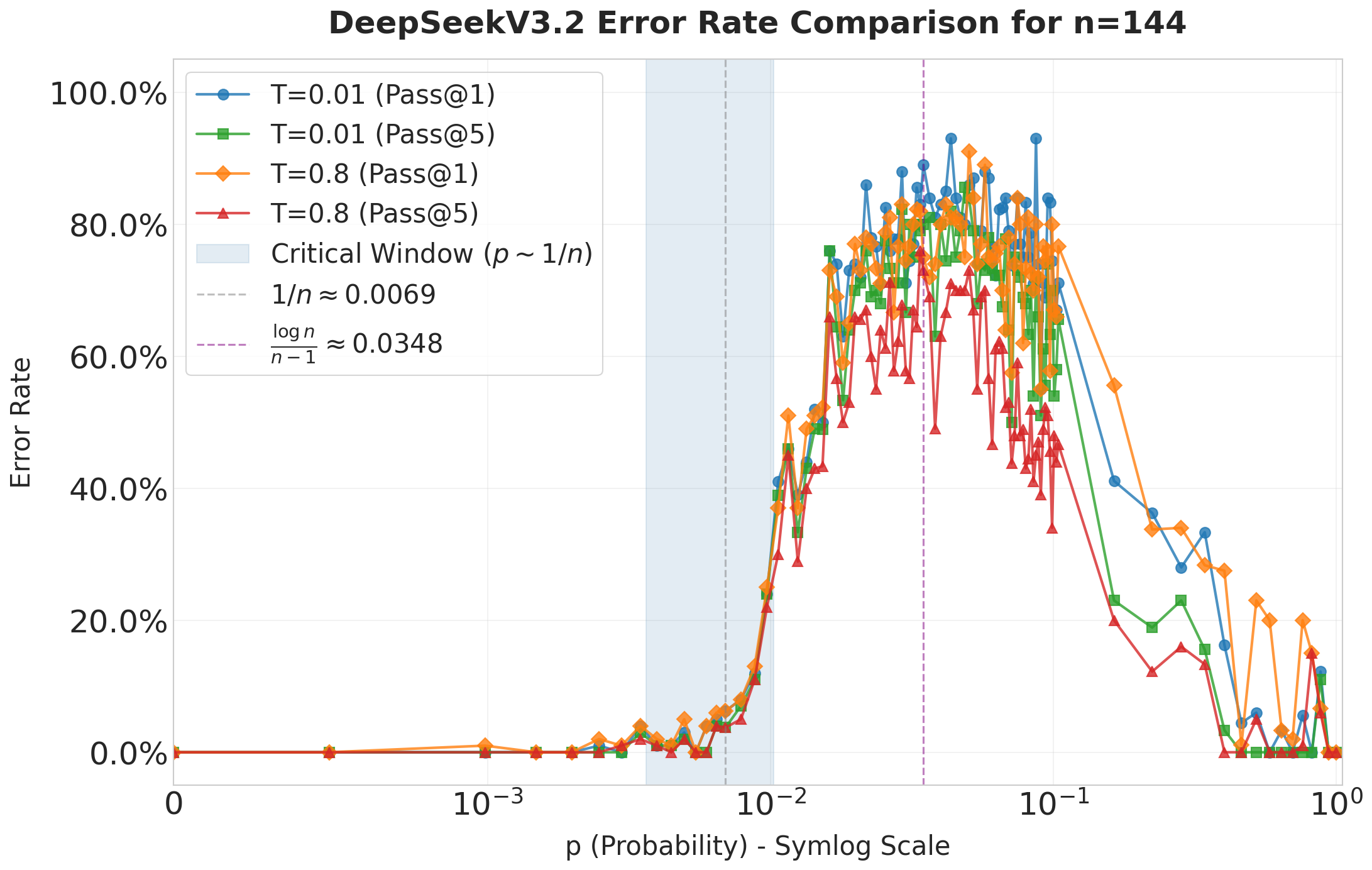}
        \caption{Decoding Strategies (Pass@K \& Temperature)}
        \label{fig:ablation_passk}
    \end{subfigure}
    
    \caption{\textbf{Ablation Studies ($n=144$).} 
    \textbf{(a)} Explicitly stating logical rules (P1) yields performance nearly identical to implicit prompts (P3).
    \textbf{(b)} Framing the problem as a Graph Theory task (P4) surprisingly outperforms the abstract Logic framing (P1).
    \textbf{(c)} Few-shot examples (P5) demonstrate comparable performance to Zero-shot (P1), indicating that ICL offers limited benefits for global logical reasoning.
    \textbf{(d)} Higher temperature ($T=0.8$) with Pass@5 (Red) improves robustness compared to greedy decoding (Blue), but cannot eliminate errors at the phase transition.}
    \label{fig:ablation_grid}
\end{figure*}

As illustrated in Figure~\ref{fig:ablation_grid}a, contrary to standard expectations, we observe \textbf{no significant performance divergence} among the three settings ($P1 \approx P2 \approx P3$). The error curves remain tightly coupled across the entire probability spectrum, responding identically to the phase transition. This yields a critical insight: DeepSeek-V3.2's failure in long-chain reasoning is not informational but structural. The model clearly ``knows" the rules of equivalence (as evidenced by its high accuracy at Depth=1), but extra definitions in the context window offer no aid when the fundamental bottleneck is the inability to execute recursive state-tracking across multiple hops. This confirms that static prompt engineering is insufficient to overcome the architectural ``One-Hop Limit".

\paragraph{2. Domain Framing: Logic vs. Graph Theory (Prompt 1 vs. 4).}
We hypothesized that the model's training corpus might be richer in algorithmic graph problems than in abstract set theory. We rephrased the task from ``equivalence classes" (Prompt 1) to ``connected components in a graph" (Prompt 4). Surprisingly, the \textbf{Graph Theory framing significantly outperformed the Logical framing} (Figure~\ref{fig:ablation_grid}b). The error rate for Prompt 4 (Brown) is consistently lower than Prompt 1 (Blue), suggesting that activating the ``graph algorithm" subspace of the model is more effective than the ``logical deduction" subspace.

\paragraph{3. The Limited Efficacy of In-Context Learning (Prompt 1 vs. 5).}
We investigated whether providing concrete examples could guide the model's reasoning process. We compared the zero-shot baseline (Prompt 1) with a few-shot setup (Prompt 5) that included explicit transitive demonstrations (e.g., ``a1=a2, a2=a3 implies a1=a3").
As illustrated in Figure~\ref{fig:ablation_grid}c, the few-shot intervention yielded results that were \textbf{statistically comparable to, or only marginally better than}, the zero-shot baseline. The error curves for Prompt 5 (Pink) and Prompt 1 (Blue) are closely entangled across the entire probability spectrum. While providing examples prevents some extreme error spikes found in the zero-shot setting, it fails to fundamentally alter the ``reasoning collapse" behavior. This suggests that while ICL (In-Context Learning) helps clarify task formatting or local rules, it is insufficient to induce the deep, recursive algorithms required for global equivalence partitioning.

\paragraph{4. Decoding Strategies (Pass@K).}
Finally, we explored whether the correct reasoning path exists in the model's latent space but is suppressed by greedy decoding. We compared Pass@1 and Pass@5 at temperatures $T=0.01$ and $T=0.8$. Results in Figure~\ref{fig:ablation_grid}d show that increasing diversity ($T=0.8$) combined with multiple samples (Pass@5) yields the best performance (Red triangle curve). However, even with Pass@5, the error rate remains high near the phase transition, indicating that sampling alone cannot fully compensate for the reasoning deficit.

\section{Conclusion}
This work establishes the Equivalence Class Problem as a controlled diagnostic probe for long-chain reasoning, revealing sharp capability boundaries in current LLMs. Three findings emerge. First, non-reasoning models collapse immediately beyond single-hop deductions, confirming that standard Transformers cannot reliably execute recursive state-tracking. Second, reasoning models exhibit a depth-dependent decay; error rates grow geometrically with chain length, indicating that CoT mechanisms expand the effective reasoning window but do not achieve true logical generalization. Third, prompt ablation suggests that these deficits stem from intrinsic architectural limitations rather than insufficient instruction.

These findings carry practical implications: deploying LLMs in domains requiring guaranteed multi-step reasoning (e.g., formal verification, legal inference) demands caution, as even state-of-the-art reasoning models exhibit systematic failure modes at scale.

Our study has limitations. ECP is deliberately homogeneous: it isolates transitive equivalence reasoning and does not cover heterogeneous real-world tasks that mix language understanding, external knowledge, planning, and tool use. Our analysis also remains behavioral rather than mechanistic. Future work should probe attention dynamics during reasoning collapse and test whether targeted fine-tuning can extend the effective reasoning depth. Ultimately, despite recent advancements, the Equivalence Class Problem remains an open challenge for Large Language Models.

\bibliographystyle{named}
\bibliography{references}

\end{document}